\pdfoutput=1

\documentclass[11pt]{article}

\usepackage[preprint]{acl}

\usepackage{times}
\usepackage{latexsym}
\usepackage[T1]{fontenc}
\usepackage[utf8]{inputenc}
\usepackage{microtype}
\usepackage{inconsolata}
\usepackage{amsmath} 
\usepackage{url}
\usepackage{booktabs}
\usepackage{amsfonts} 
\usepackage{nicefrac}
\usepackage{microtype}
\usepackage{xcolor}   
\usepackage{multirow}
\usepackage{algorithmic}
\usepackage{graphicx}
\usepackage{textcomp}
\usepackage{comment}
\usepackage{color, colortbl}
\definecolor{Gray}{gray}{0.9}
\usepackage{tabularx}
\usepackage{arydshln}
\usepackage{tikz}
\usepackage{pgfplots}
\usepackage{tcolorbox}
\usepackage{graphicx}
\usepackage[export]{adjustbox}
\usepackage{subcaption}
\usepackage{enumitem}
\usepackage{xr}
\usepackage[normalem]{ulem}
\usepackage{hyperref}
\usepackage{cleveref}
\newcommand{\Bench}{\textsc{LEGOBench}}

\title{\Bench: Scientific Leaderboard Generation Benchmark}

\author{Shruti Singh$^*$ \and Shoaib Alam$^*$ \and Husain Malwat \and Mayank Singh \\
LINGO, Indian Institute of Technology Gandhinagar, India \\
\{singh\_shruti, shoaibalam, husainmalwat, singh.mayank\}@iitgn.ac.in}

\begin{document}
\maketitle
\def\thefootnote{*}\footnotetext{Equal contribution}\def\thefootnote{\arabic{footnote}}

\begin{abstract}
The ever-increasing volume of paper submissions makes it difficult to stay informed about the latest state-of-the-art research. To address this challenge, we introduce \Bench, a benchmark for evaluating systems that generate scientific leaderboards. \Bench~is curated from 22 years of preprint submission data on arXiv and more than 11k machine learning leaderboards on the PapersWithCode portal. We present four graph-based and two language model-based leaderboard generation task configurations. We evaluate popular encoder-only scientific language models as well as decoder-only large language models across these task configurations. 
State-of-the-art models showcase significant performance gaps in automatic leaderboard generation on \Bench.
The code is available on GitHub\footnote{\url{https://github.com/lingo-iitgn/LEGOBench}} and the dataset is hosted on OSF\footnote{\label{osflink}\url{https://osf.io/9v2py/?view\_only=6f91b0b510df498ba01595f8f278f94c}}.
\end{abstract}

\section{Introduction}
Comparison of results with prior state-of-the-art (SOTA) is a standard practice in experimental research papers. Performance on a task using a specific metric establishes the efficacy of the paper's proposed method. However, one of the primary challenges in scientific research is keeping up with the rapid volume of research progress and staying updated with the latest SOTA to compare with one's work.
The increasing number of manuscripts (depicted by arXiv submissions in~\Cref{ssec:app_expgrowth}) showcases the severity of information overload. With the continuous stream of submission, revision, and acceptance timelines of conferences and journals, researchers often struggle to keep up with the latest methods and developments. Thus, being acquainted with the latest papers, sieving through the massive set, and deciding which baselines to compare with can be challenging and time-consuming. Moreover, the latest papers with novel methods may have low visibility, and upcoming papers may overlook those for result comparison as citations are biased towards old compared to new papers.

To address the information overload and to facilitate the comparison with meaningful baseline works, multiple previous works mine scientific tables from papers~\citep{kayal2022tables, tabimghtmlpubtabnet2020,deng2019challenges,liu2007tableseer} and construct scientific leaderboards~\citep{yang2022telin, kabongo2023zero, kabongo2021automated, kardas2020axcell, hou2019identification}. A scientific leaderboard curates performance scores of competitive models against the triple <dataset, task, metric>. One of the most actively maintained platforms, PapersWithCode (PwC)~\citep{pwc}, hosts leaderboards in empirical machine learning.~\Cref{fig:rep_leaderboards} in~\Cref{ssec:rep_ldb_pwc} shows a representative leaderboard sample for the image clustering task on the MNIST dataset, available in PwC. 
% The competing models are sorted based on their performance against the NMI metric. 
While leaderboards are helpful for researchers to track the latest models, a majority of leaderboard curation initiatives are manually maintained~\citep{pwc}, or are dormant~\citep{effaimetrics, redditsota, nlpprogress}. Hence, the need to automate the generation of leaderboards is imperative.

\begin{figure*}[!htbp]
    \centering
    \includegraphics[width=0.8\textwidth]{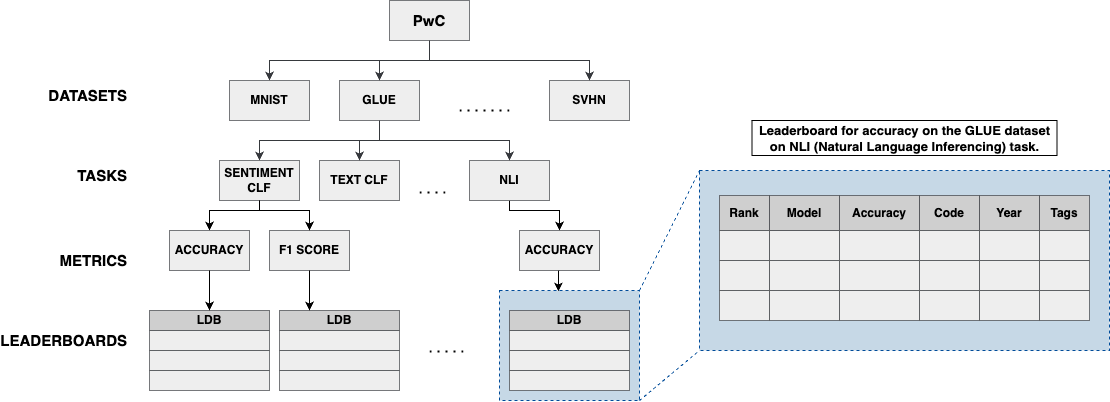}
    \caption{Organization of leaderboards in PwC. A leaderboard is constructed for a <dataset, task, metric> tuple. Leaderboards can contain additional metadata, such as the code repository link and model description tags.}
    \label{fig:pwc_hierarchy}
\end{figure*}

To streamline the process of automating leaderboard generation, we create the arXiv Papers' Collection (APC), a curated collection of research papers and graph data (citation network and performance comparison network) from arXiv. We also create a dataset sourced from PapersWithCode (PwC), consisting of leaderboards mapped with arXiv papers. We combine APC and PwC datasets to develop a benchmark framework called \Bench, that facilitates evaluation and assessment of automatic leaderboard generation models. \Bench~introduces the leaderboard generation task in three configurations, (i) Ranking Papers based on Content and Graph [RPG], (ii) Ranking Papers by Prompting Language Models [RPLM], and (iii) Leaderboard Entries Generation by Prompting Language Models [LGPLM]. Our contributions can be summarized as follows:
\begin{itemize}[noitemsep,nolistsep]
    \item We curate the first leaderboard generation framework, \Bench, where we provide datasets and metrics for evaluating scientific leaderboard generation. Our dataset consists of 22 years of arXiv data and 11k leaderboards from PwC, available publicly on OSF\footref{osflink}.
    \item We present six leaderboard generation task configurations, including four that are graph-based and two that utilize language models. The diverse task configurations allow for a comprehensive evaluation of systems, showcasing our framework's adaptability and breadth across differing methodologies.
    \item We assess the ability of the existing off-the-shelf encoder-only scientific LMs and decoder-only LLMs in the context of leaderboard generation. Our results showcase the severe limitations of existing models, uncovering avenues for future models to address.
\end{itemize}

 % We discuss the curation process of APC and PwC datasets and their linkage in~\Cref{sec:dataset}. We then discuss the \Bench~task configurations in~\Cref{sec:benchmark_details}. Preliminary baselines and results follow in~\Cref{sec:prelimbaselines}. We discuss the previous attempts at leaderboard generation in~\Cref{sec:relatedworks}. Finally, we conclude with the insights from our experiments on leaderboard generation along with the potential usage of \Bench~in~\Cref{sec:conclusion}.

\section{Dataset}
\label{sec:dataset}
We curate two datasets, (i) PwC Leaderboards (PwC-LDB) and (ii) arXiv Papers' Collection (APC), which are utilized for the construction of \Bench. PwC-LDB is curated from Papers With Code repository\footnote{\label{fn1pwc}https://paperswithcode.com/. Dataset curation - 06/2023.}  and APC is curated from arXiv preprint repository\footnote{\label{fn2ax}https://arxiv.org/. Dataset curation - 09/2022.}. %This section describes the curation methodology of the PwC-LDB and APC datasets. 

\begin{table}[!htbp]
    \centering
    \small{
        \begin{tabular}{lrr}
        \hline
        \rowcolor{Gray}
        \textbf{Artifact} & \textbf{PwC-LDB} & \textbf{AP-LDB} \\  \hline
        Datasets & 3666 & 1697     \\
        Tasks & 1660 &   675    \\ 
        Metrics & 2958 & 1381      \\
        DTMA & 70559 & 43105 \\
        Leaderboards & 11470 & 4409 \\
        \hline
        \end{tabular}
    }
    \caption{Statistics of PwC-LDB \& AP-LDB dataset. DTMA refers to <data, task, metric, method> tuple representing an entry in the leaderboard. AP-LDB is curated by mapping PwC-LDB with APC.}
    \label{tab:pwcldb_stats}
\end{table}

\begin{figure*}[!tbp]
    \centering
    \includegraphics[width=0.7\textwidth]{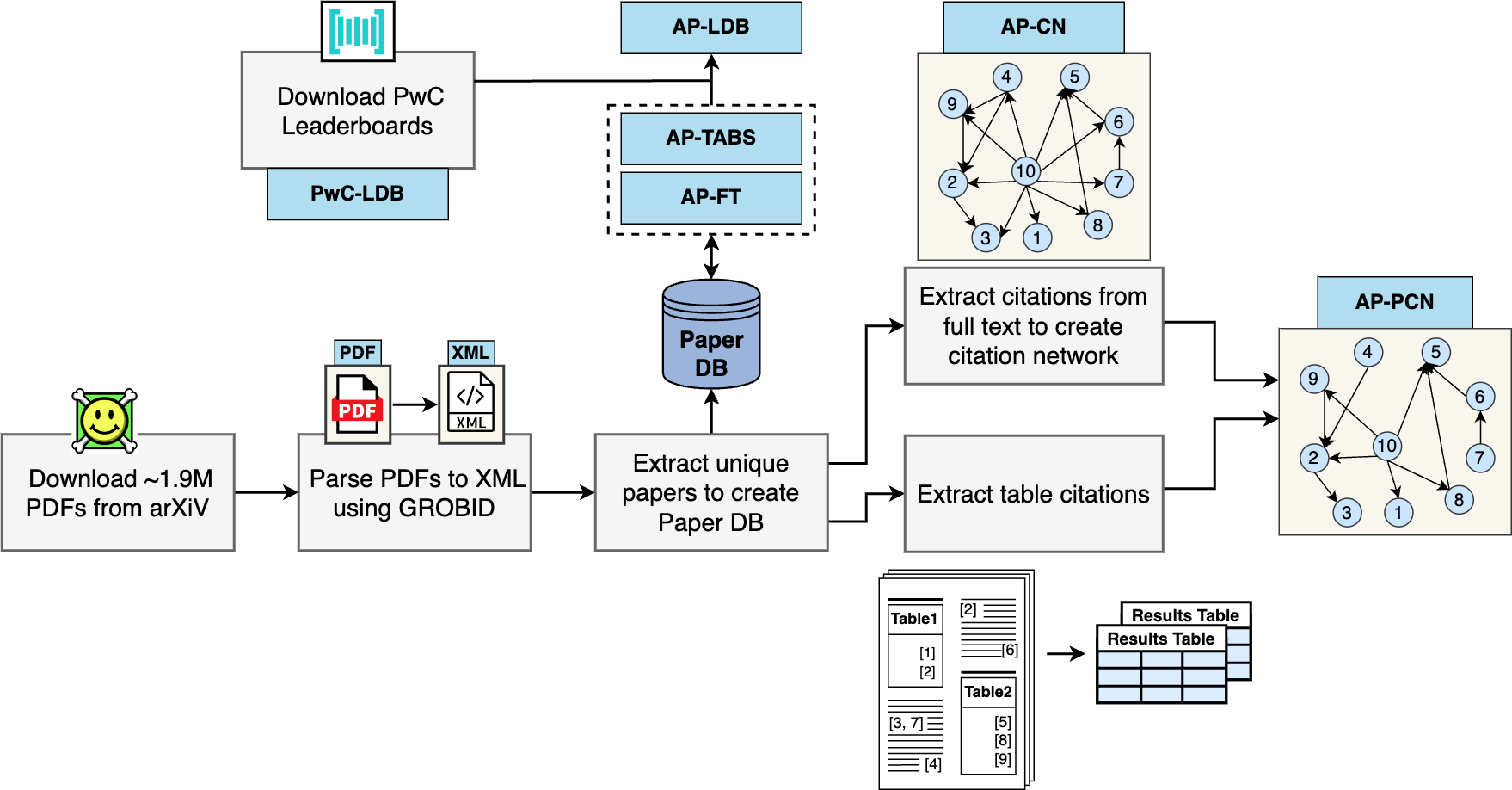}
    \caption{Pipeline for constructing the APC datasets. Blue boxes denote various datasets in the APC collection and the PwC-LDB dataset.}
    \label{fig:datacurationpipeline}
\end{figure*}

\subsection{PwC-LDB}
\label{ssec:data_pwcldb}
PwC-LDB is a dataset of leaderboards for various ML tasks curated by the PwC repository. The PwC repository is curated and updated regularly. It is organized in a hierarchy, with ML datasets forming the parent nodes and tasks associated with dataset nodes being the children. Each task is evaluated using certain metrics, and hence each leaderboard is associated with a dataset (D), task (T), and metric (M), as represented in~\Cref{fig:pwc_hierarchy}. Formally, a leaderboard $\mathcal{L}$(D, T, M) is defined for a triplet <D, T, M>, where an algorithm/method/model (hereafter denoted as A) is evaluated against D, T, and M. <D, T, M, A>, thus, represents the addition of algorithm/method/model to the DTM triple. Every leaderboard consists of multiple A's that compare their performance scores against each other. Information such as code repository, year, model parameters, etc., could also be present.

We curated 3666 machine learning (ML) datasets, 1660 tasks, and their corresponding leaderboards from the PwC repository. 11,470 leaderboards contain 70,559 <D, T, M, A> tuples. On average, a leaderboard contains six entries, i.e., a performance comparison of six models. The maximum number of entries in a leaderboard is 863 for the `Image classification on ImageNet using Top-1 Accuracy' leaderboard. The statistics of the PwC-LDB dataset are presented in~\Cref{tab:pwcldb_stats}.

\subsection{APC}
\label{ssec:data_atc}
\begin{table*}[!t]
    \centering
    \small{
    \begin{tabular}{cp{0.55\textwidth}p{0.25\textwidth}}
    \hline
    \rowcolor{Gray}
    \textbf{Dataset} & \textbf{Summary} & \textbf{Size} \\ \hline
        AP-TABS & Titles and Abstract of arXiv papers extracted from metadata & 1.9M Papers\\
        AP-FT   & Full text of arXiv papers extracted after parsing PDFs with GROBID & 1.9M Papers\\
        AP-CN   & Citation network of arXiv papers extracted using regex from AP-FT & 18M nodes \& 59M edges\\
        AP-PCN  & Performance comparison network extracted from table citations using AP-FT and AP-CN & 280k nodes \& 309k edges\\
        AP-LDB  & PWC-LDB mapped with the arXiv Papers' Collection & 9.8k leaderboards \& 41k <DTM> \\ \hline
    \end{tabular}
    }
    \caption{Summary of datasets in the \Bench~benchmark. %AP prefix denotes the arXiv datasets curated specifically in this work.
    }
    \label{tab:data_summary}
\end{table*}

APC is a collection of datasets curating diverse paper information from arXiv\footref{fn2ax}, a research-sharing platform that hosts preprints of scientific papers in eight domains. We curate titles and abstracts (AP-TABS), full-texts (AP-FT) from arXiv, and process the data to extract the citations (AP-CN) and performance comparisons (AP-PCN). Next, we discuss the stages (denoted by [S\emph{i}]) in the APC curation pipeline. ~\Cref{fig:datacurationpipeline} illustrates the pipeline. %The datasets in APC extracted during each of the stages in the pipeline are also discussed. 
\\
\underline{\textbf{[S1] arXiv Paper Curation:}} We curate the arXiv data by collecting metadata and paper PDFs from Jan 2000 to July 2022, consisting of 1,942,301 papers categorized into eight broad domains. The domain-wise statistics of the curated papers are presented in~\Cref{ssec:app_networkstats}. %Out of 1,942,301 papers, the full text of 1,940,910 papers is present. 
In the remainder of this paper, we refer to the title and abstract metadata obtained directly as the \textbf{AP-TABS} dataset (arXiv Papers' Titles and Abstracts).\\
\underline{\textbf{[S2] Parsing PDFs:}} We parse PDFs into TEI-XML format using GROBID~\citep{lopez2009grobid}. GROBID successfully extracts full-text information from 1,940,910 papers, which is referred to as \textbf{AP-FT} (arXiv Papers' Full Text) dataset. \\
\underline{\textbf{[S3] Constructing the Unique Paper Index:}} We construct a unique index of all papers present in our dataset. This index includes 1.9M papers present in AP-FT along with their references.\\
\underline{\textbf{[S4] Construction of the %arXiv
Citation \textbf{Network:}}} The arXiv Papers' Citation Network dataset (\textbf{AP-CN}) consists of citations in the AP-FT dataset (details in~\Cref{ssec:app_networkstats}). 
However, the leaderboards in the PwC-LDB dataset are present only for computer science, specifically in ML.\\
\underline{\textbf{[S5] Table Extraction and Construction of the}} 
\underline{\textbf{Performance Comparison Network:}} Finally, we extract tabular information from parsed TEI-XML paper format in the AP-FT dataset. Citations in the table are extracted and mapped to the papers in the unique index. It should be noted that the table citation data of a paper is a subset of the references of the paper. We only include papers containing at least one table with at least one citation in the table text or the caption to construct the \textbf{AP-PCN} (arXiv Papers' Performance Comparison Network) data. \\
\underline{\textbf{[S6] Mapping PwC-LDB with APC:}} \label{ssec:mappwcapd}
PwC-LDB dataset consists of leaderboards for a task, dataset, and metric triple, denoted by <T, D, M>. Each model A is listed with the title of the paper and its unique arXiv Identifier, if available. We leverage these identifiers to map to the AP-FT dataset. 
The paper's metadata or full-text information is originally absent in the PwC-LDB, which we collate and map in the AP-LDB dataset.\\ 
\noindent A summary and statistics of datasets in \Bench~are presented in~\Cref{tab:data_summary} and size, domain, and modality statistics of the AP-CN and the AP-PCN dataset are presented in~\Cref{ssec:app_networkstats}.

\section{\Bench: Automatic Scientific Leaderboard Generation Benchmark }
\label{sec:benchmark_details}
We present \Bench, a benchmark specifically developed for the evaluation of scientific leaderboard generation. This benchmark tasks systems with generating a leaderboard in response to a natural language query that specifies a dataset (D), a task (T), and a metric (M), utilizing a collection of arXiv Papers' datasets. To comprehensively assess the capabilities of scientific leaderboard generation systems, we design three tasks, resulting in a total of six configurations. These configurations employ multiple frameworks and use various datasets from the arXiv Papers' Collection to assess the generation of diverse leaderboard formats.

The three leaderboard generation tasks are: (i) Ranking Papers based on Content and Graph [RPG], (ii) Ranking Papers by Prompting Language Models [RPLM], and (iii) Leaderboard Entries Generation by Prompting Language Models [LGPLM]. The benchmark for the first two ranking tasks comprises 4409 leaderboards for 675 empirical ML tasks on 1697 datasets. The LGPLM task comprises 9847 leaderboards from the AP-LDB dataset. For each task, the format is:
Given query \emph{q} = <D, T, M>, and arXiv dataset $\mathcal{D}$, the task is to generate a leaderboard $\mathcal{L}$ corresponding to the query \emph{q}.
The format of \emph{q}, $\mathcal{D}$ and $\mathcal{L}$ depend on the task configuration. Next, we describe each task in detail and discuss the four task configurations for the RPG task and one configuration for the RPLM and LGPLM tasks. 

\subsection{Ranking Papers based on Content and Graph [RPG]}
Given a short natural language query \emph{q} (consisting of D, T, and M details), this task format requires ranking candidate papers based on the performance score. For RPG tasks, $\mathcal{L}$ is a ranked list of papers, where the best rank indicates the best performance on the <D, T, M> triple. The first step is retrieving a set of candidates from the arXiv Papers' Collection. It leverages the network structure as well as the paper content, to generate a ranked list of papers such that the papers with the best performance are ranked highest, and ranks increase as performance decreases. We encourage the evaluation of graph models for this task format as we present multiple configurations of this task with different text and network datasets. 
\begin{enumerate}[noitemsep,nolistsep]
    \item \textbf{Ranking Papers in the Citation Network with Titles and Abstracts (RPG[CN-TABS]):} Given the title and abstract of each paper in the arXiv Papers' Citation Network, i.e., $\mathcal{D}$ = AP-CN $\bigcup$ AP-TABS, construct the leaderboard based on the citation network properties and the content.
    \item \textbf{Ranking Papers in the Performance Comparison Network with Titles and Abstracts (RPG[PCN-TABS]):} Given the title and abstract of each paper in the arXiv Papers' Performance Comparison Network, i.e., $\mathcal{D}$ = AP-PCN $\bigcup$ AP-TABS, construct the leaderboard based on the comparison network properties and the content present in TABS. The AP-PCN dataset encodes which paper compares results with which papers; however, if the performance is better or worse, it is not present in the graph dataset. The improvement needs to be extracted from the TABS dataset and used in addition to the comparison data to generate the correct ranking of papers in the leaderboard.
    \item \textbf{Ranking Papers in the Citation Network with Full Texts (RPG[CN-FT]):} Given access to the full text of papers along with the citation network, i.e. $\mathcal{D}$ = AP-CN $\bigcup$ AP-FT, this task focuses on generating the ranked paper list by leveraging the network as well as the full text. For example, the full text can be utilized to learn node embeddings in the graph.
    \item \textbf{Ranking Papers in the Performance Comparison Network with Full Texts (RPG[PCN-FT]):} This task is similar to RPG[CN-FT], except that instead of the citation network, performance comparison network is provided. $\mathcal{D}$ = AP-PCN $\bigcup$ AP-FT. 
\end{enumerate}
% TODO: We posit that the full-text data vs tabs comparison (bring from other places if text available, ideally should be), cn vs pcn discussion

\subsection{Ranking Papers by Prompting Language Models [RPLM]}
RPLM focuses on ranking the papers using language models (LMs). Given a natural language query \emph{q} (consisting of D, T, and M details), and $\mathcal{D}$ is a randomly shuffled list of paper titles present in the leaderboard corresponding to <D, T, M>. LMs are expected to generate output $\mathcal{L}$ as a ranked list of paper titles, such that the best-ranked paper in the list achieves the best score on <D, T, M>. This task intends to leverage LMs to rank papers by retrieving the best performance score of the model discussed in the paper.
This task opens the possibilities for evaluating if a language model encodes/memorizes relevant information about the paper results in the model parameters as paper text is not provided.
% The task also examines the capabilities of LM to infer the information present in scientific tables and charts. RPLM relies on language models to extract scores from memory and generate ranks.
However, it is a challenge to automatically ascertain whether the rationale behind the generated rankings truly takes into account the extracted scores. We use the AP-LDB dataset to construct the queries. As this is a ranking task, we only use leaderboards where at least three unique papers are present so that it is practical to evaluate the generated paper title rankings.

\subsection{Leaderboard Entries Generation by Prompting Language Models [LGPLM]}
LGPLM is modeled as a QA task over documents, where given the query \emph{q} (consisting of D, T, and M details), and $\mathcal{D}$ = AP-FT, a language model extracts method performance from papers experimenting on T and D and reporting scores with M. It focuses on the extraction of leaderboard entries, consisting of method/model/algorithm names (A in <D, T, M, A> tuple, and referred as method hereafter in the paper) and the performance scores for metric M and arranging them into a leaderboard. The $i^{th}$ leaderboard entry in $\mathcal{L}$ can be represented as <$m_i$, $s_i$>, where $m_i$ is the method name and $s_i$ is the score. In our dataset, $\mathcal{L}$ is stored as a markdown table containing <$m_i$, $s_i$> entries, in string format. We chose the markdown table format for storing the leaderboard entries as our preliminary evaluation highlighted that most LLMs are efficient at generating a uniform markdown table rather than any other format (e.g. tab or space separated columns). However, during evaluation, we do not use straightforward rouge-based evaluations to compare the LLM output with the ground truth. Instead, we parse the string to extract methods and scores. Metrics are discussed in~\Cref{ssec:task_eval}.

The three tasks, RPG, RPLM, and LGPLM are designed for different purposes. While the RPG task focuses on paper graph representation models and models for ranking graph nodes, the RPLM task focuses on retrieving the score information from the memory of scientific language models. LGPLM focuses on the extraction of methods and their scores from the paper text and utilizing that for leaderboard generation.

\begin{comment}
We summarize the task configurations for the leaderboard generation task for RPG and SRLM in~\Cref{tab:rpg_srlm_taskconfigs}.
% and for SRLM in~\Cref{tab:srlm_taskconfigs}.

\begin{table}[!tbph]
\centering
\caption{RPG and SRLM Task Configurations.}
\begin{tabular}{llll}
\hline
\rowcolor{Gray}
\textbf{Task Config} & \textbf{Datasets $\mathcal{D}$} & \textbf{Content}  & \textbf{Network}   \\ \hline
    % RPG[CN]         & AP-CN             & --                & Citation Network                 \\
    % RPG[PCN]        & AP-PCN            & --                &  Performance Comparison Network  \\ 
    RPG[CN-TABS]     & AP-CN + AP-TABS    & Title \& Abstract & Citation Network                 \\
    RPG[PCN-TABS]    & AP-PCN + AP-TABS   & Title \& Abstract & Performance Comparison Network   \\ 
    RPG[CN-FT]       & AP-CN + AP-FT      & Full text         & Citation Network                 \\
    RPG[PCN-FT]      & AP-PCN + AP-FT     & Full text         & Performance Comparison Network   \\ \hline
\end{tabular}
\label{tab:rpg_srlm_taskconfigs}
\end{table}
\end{comment}

\subsection{Evaluation of Leaderboard Generation}
\label{ssec:task_eval}
For the RPG task, we use Kendall's Tau (KTau)~\citep{kendall1938new} to measure the rank correlation between the ranked list of paper titles generated by the candidate model and the ground truth ranks of papers in the leaderboard. It is in the range of -1 to +1, indicating perfect inverse or direct association or no association if 0. We also describe some custom-designed metrics. \textbf{Binary Exact Match (BEM)} takes binary values, i.e., one iff the two ranked lists are exactly similar; otherwise zero. To enhance readability, we present BEM percentage values, i.e. the percentage of ranked lists that are ordered. \textbf{Complete Inclusion Score (CIS)} CIS denotes the percentage of model-generated leaderboards that have all the titles present in the ground truth. \textbf{Concordant Pairs (CP)} measures the percentage of pairs of leaderboard entries ranked in the same order as in the ground truth. It lies in the range 0-100, with 100\% indicating that all pairs are ranked in the same order. \textbf{Method Recall (MR)} is used for the LGPLM task. The output of the LGPLM task is a leaderboard consisting of method/algorithm names and their corresponding scores, arranged in a markdown table. This metric computes the percentage of correct method names in the model-generated leaderboard with respect to the method names in the ground truth. \textbf{Method Precision (MP)} is similar to MR, and computes the percentage of correct methods in the model-generated leaderboard with respect to the total number of generated methods. \textbf{Score Precision (SP)} computes the percentage of correct scores in the model-generated leaderboard with respect to the total number of generated methods.

\section{Preliminary Baselines and Results}
\label{sec:prelimbaselines}
We present preliminary baselines for each of the tasks. Due to space constraints, the schematic diagrams for the baselines are presented in~\Cref{app:baseline_diagrams}.

\begin{table}[!htbp]
\centering
\small{
\begin{tabular}{c|cc|cc}
\rowcolor{Gray}
Task $\rightarrow$ & \multicolumn{2}{c|}{\textbf{RPG{[}CN-TABS{]}}} & \multicolumn{2}{c}{\textbf{RPG{[}PCN-TABS{]}}} \\
\rowcolor{Gray}
Model $\downarrow$  & \textbf{BEM} & \textbf{KTau} & \textbf{BEM}  & \textbf{KTau} \\ \hline
SciBERT  & 2.66  & -0.010 &  8.26  & -0.187   \\
SPECTER  & 2.60  & -0.009 &  8.74  & -0.177   \\
SciNCL   & 2.25 & -0.015 & 8.51  &  -0.175   \\
OAG-BERT & 2.62 & -0.009 & 8.34  & -0.201    \\
\end{tabular} 
}
\\
\vspace{3ex}
\small{
\begin{tabular}{c|cc|cc}
\rowcolor{Gray}
Task $\rightarrow$ & \multicolumn{2}{c|}{\textbf{RPG{[}CN-FT{]}}}  & \multicolumn{2}{c}{\textbf{RPG{[}PCN-FT{]}}} \\
\rowcolor{Gray}
Model $\downarrow$ & \textbf{BEM} & \textbf{KTau} & \textbf{BEM} & \textbf{KTau}  \\ \hline
SciBERT  & 0.184 & -0.006 & 6.743 & -0.140   \\
SPECTER  & 0.851 & -0.008 & 6.978 & -0.137   \\
SciNCL   & 0.888 & -0.006 & 6.283 & -0.015   \\
OAG-BERT & 0.665 & -0.010 & 7.201 & -0.132   \\       
\end{tabular}
}
\caption{Performance of PageRank for ranking nodes. Candidates Retrieval selected intersecting candidate documents retrieved by query unigram search.}
\label{tab:rpg_task_performance}
\end{table}

\subsection{RPG Baselines and Results}
We follow a retrieve-then-rank procedure for ranking papers based on content and network as depicted in~\Cref{fig:rpg}. The first step is candidate retrieval, which selects a subset of papers from the arXiv dataset, followed by a ranker module that ranks the papers to generate leaderboard entries.

\noindent\textbf{Candidate Retrieval Module:} We present a straightforward methodology to retrieve candidate papers given the query. We tokenize the query and obtain unigrams. The query unigrams are searched in papers (title and abstract if AP-TABS, and full-text for AP-FT configuration), and the papers containing all the unigrams are selected. 
% We calculate recall to measure relevant papers retrieved via this unigram retrieval module. 
% We also experimented with bigrams; however, the performance deteriorated. 
% In another experiment, we retrieved the papers which contained at least one of the keywords from the query, which is similar to taking a union of all candidates containing atleast one unigram from the query. We present the results for the union and intersection of ngram retrieval for unigrams and bigrams in~\Cref{tab:recall_candidate_retrieval}. As unigram retrieval and candidate union retrieved \textcolor{blue}{todo} papers out of \textcolor{blue}{todo} papers in the dataset, we decided to use candidate intersection instead of union.
% \begin{table}[h]
% \centering
% \caption{Recall for unigram and bigram retrieval for fetching candidates. We only considered leaderboards that contain atleast \textcolor{blue}{three} papers from the arXiv papers dataset before June 2022}
% \label{tab:recall_candidate_retrieval}
% \begin{tabular}{c|cc}
% \hline
% \rowcolor{Gray}
%  & \multicolumn{2}{c|}{\textbf{Recall - Candidate Intersection}} \\
% \rowcolor{Gray}
% \multirow{-2}{*}{\textbf{Parameter}} & AP-TABS & AP-FT \\
% \hline
% Mean & 64.77 & 27.09 \\
% Median & 66.67 & 12.04  \\
% Mode & 60.0 & 0.0 \\
% SD ($\sigma$) & 23.36 & 32.87 \\ \hline
% % Max & 100.0 & 100.0 & 100.0 & 100.0 \\ \hline
% \end{tabular}
% \end{table}

\noindent\textbf{Ranker Module:} After candidate retrieval, we construct a graph of the retrieved papers.
The candidate papers retrieved in the previous step are considered nodes, and directed unweighted edges are added from the citation or performance comparison network depending on the dataset provided with the task. Weights are added to the existing edges by encoding paper content (TABS) and computing cosine similarity between the paper nodes. 
% The cosine similarity value between two papers constitutes the weighted edge if an edge originally existed between the pair in the network. 
We experimented with popular scientific encoder models such as SciBERT~\citep{beltagy-etal-2019-scibert}, SPECTER~\citep{cohan-etal-2020-specter}, SciNCL~\citep{ostendorff-etal-2022-neighborhood}, and OAG-BERT~\citep{liu2021oag}, to encode the paper content. Node scores are calculated using PageRank~\citep{Page1999ThePC} and are used to rank the nodes. 
We report the leaderboard generation performance results in~\Cref{tab:rpg_task_performance}. KTau values close to 0 for all baselines indicate that there is no correlation between the generated lists. The maximum BEM values are in the range of 6-8\%, indicating less than random chance papers being ranked correctly. Among the baselines, the usage of AP-PCN shows promise over AP-CN as performance comparison is a substantial signal compared to citations. In the current settings, usage of the AP-FT leads to inferior performance than AP-TABS, however, we posit the large candidate set retrieved from AP-FT to be the actual reason. AP-FT usage could potentially enhance performance as it is more likely to discuss performance scores exhaustively in the full-text, contingent upon the integration of a more effective candidate retrieval module. Performance comparison networks and paper full-text datasets have long been unexplored and have the potential to improve leaderboard generation performance. Existing baselines however perform poorly, thereby presenting an opportunity to leverage the benchmark for the assessment of novel graph and encoder LMs.

\subsection{RPLM Baselines and Results}
For the RPLM task, we prompt language models by adding an instruction to the provided natural language query \emph{q} and paper titles. The pipeline and complete prompt are presented in~\Cref{fig:rplm} and~\Cref{tab:rplm_prompt} respectively. We present results with open-source LLMs (Falcon~\citep{falconmodel}, Galactica~\citep{taylor2022galactica}, LLama 2~\citep{llama2model}, Mistral~\citep{jiang2023mistral}, Vicuna~\citep{vicuna2023}, and Zephyr~\citep{zephyrmodel}) and two closed models Gemini~\citep{team2023gemini} and GPT-4~\citep{achiam2023gpt}. We evaluate 7B and 13B models and exclude bigger models due to resource constraints.

Results for the models are presented in~\Cref{tab:srlm_prompting}. 
LLama 2 Chat 7B and Vicuna 7B perform best among the open-source LLMs in generating a ranked list that consists of all the titles provided in the input prompt (i.e. CIS metric) for around 16\% of the instances in the dataset. The 7B versions of Llama 2 and Vicuna perform better at generating the same titles (and hence understanding the instruction correctly) than their counterpart 13B models. Overall, GPT-4 has the highest CIS of 21.89\%.
BEM and KTau are calculated for the subset of instances for which the LLM-generated ranked list contains all the papers in the ground truth, hence we omit these values for papers with CIS less than 1\%. BEM values are less than 1\% for all models, indicating less than 1\% of the LLM-generated ranked paper titles are exactly the same as the original ranking of papers. Similarly, all models' KTau values are close to zero, indicating no association between the LLM-generated paper ranks and the original leaderboard ranks. Finally, we present CP, indicating the percentage of concordant pairs in the LLM-generated titles and original leaderboard titles. Note that LLM-generated titles absent from the original prompt input are ignored while computing CP. GPT-4 has the highest CP with 51.93\% concordant title pairs, followed by Gemini with 46.06\% concordant title pairs. The best performing open-source LLMs, Llama 2 Chat 7B and Vicuna 7B lag behind GPT-4 with around 16 points. Instruction-tuned models perform better than their regular counterparts across all the open-source LLMs.

Manual analysis of generated ranks revealed that Gemini follows the prompt instruction efficiently by generating a well-formatted ranked paper list. Most open-source models such as Falcon, Llama, Vicuna, and Mistral generate titles absent in the prompt and often also repeatedly generate the same title. 
Manual inspection of the results indicates that most models are unable to follow the instructions and often end up generating paper titles that were not provided for ranking in the input. These titles are often hallucinated and no papers with such titles exist in the public domain. Falcon 7B, Llama 2 7B, and Llama 2 13B often keep generating the same paper title multiple times in the ranked list, however, the instruction-tuned counterparts of these models generally did not face this issue. The majority of the Galactica-generated answers have HTML tags (specifically <s>, <p>, and sometimes <li>), or the first generated title is repeated in the entire generated text. Galactica model also often ends up generating long text in the format of a paper title and abstract instead of paper title ranks. We also observe that the Vicuna 13B model is chattier in comparison to the Vicuna 7B model despite the instruction clearly stating to only generate the ranked titles and skip any additional text. We posit this as the reason for the slightly better performance of the Vicuna 7B model compared to the 13B model. 

\begin{table}[!htbp]
    \centering
    \small{
    \begin{tabular}{lcccc}
    \hline
    \rowcolor{Gray}
    \textbf{Model}  & \textbf{CIS} & \textbf{BEM} & \textbf{KTau} & \textbf{CP} \\ \hline
    \multicolumn{5}{c}{7 B Models}      \\ \hline
    Falcon & 1.24 & 0.20 & -0.17 & 1.65 \\
    Falcon Instruct & 7.46 & 0.23 & 0.00 & 16.38 \\
    Galactica & 0.50 & - & - & 0.00 \\ 
    LLama 2 & 0.25 & - & - & 1.36 \\
    LLama 2 Chat & 16.17 & 0.11 & 0.02 & 35.87 \\
    Mistral & 1.74 & 0.29 & -0.04 & 3.25 \\
    Mistral Instruct & 3.48 & 0.07 & -0.05 & 7.97 \\
    Vicuna & 16.92 & 0.10 & 0.05 & 36.87 \\
    Zephyr Beta & 7.21 & 0.10 & -0.14 & 13.25 \\ \hline
    \multicolumn{5}{c}{13 B Models}      \\ \hline
    LLama 2 & 0.25 & - & - & 0.46 \\
    LLama 2 Chat & 10.20 & 0.07 & 0.01 & 29.71 \\
    Vicuna & 14.18 & 0.09 & 0.07 & 34.66 \\ \hline
    \multicolumn{5}{c}{Closed Models}      \\ \hline
    Gemini Pro & 18.91 & 0.09 & 0.06 & 46.06 \\
    GPT-4 & 21.89 & 0.08 & 0.10 & 51.93 \\
    \hline
    \end{tabular}
    }
    \caption{Performance of LLMs on the RPLM task.}
    \label{tab:srlm_prompting}
\end{table}

\subsection{LGPLM Baselines and Results}
We follow a retrieval-augmented-generation setup for the LGPLM baseline. We use a BM25 ranker module that takes the leaderboard query and full-text paper chunks as input and selects top-k chunks. For our experiments, only the papers present in the leaderboard (i.e. papers that report performance on the specified task and dataset using the specified metric) are split into chunks and provided to the BM25 ranker. We use k=10 for our experiments. The top-k chunks and the query are then provided to a language model, which generates the leaderboard consisting of methods and performance scores. We include the top-10 chunks iteratively in the prompt and keep including the chunks till the model context length is exhausted. The pipeline is presented in~\Cref{fig:lgplm} and prompt details in~\Cref{tab:lgplm_prompt}. We use the same set of LLMs as used in the RPLM task.

We present the results for this configuration in~\Cref{tab:lgplm_results}. The Method Recall (MR) is less than 1\% for 10 out of the 14 evaluated models (Falcon 7b, Falcon Instruct 7B, Galactica 6.7B, Llama 2, Llama 2 Chat 7B, Llama 2 13B, Llama 2 Chat 13B, Vicuna 7B, and Vicuna 13B). 
% All these models ended up generating only noisy table headers (some variation of "method" and metric) for the leaderboard. The poor MR scores with less than 1\% values for the 10 models render the computation of score precision impractical. 
A manual examination of the results reveals that the majority of models merely produce the leaderboard table header in the format "Method | Metric," thereby precluding the possibility of assessing Score Precision (SP). Further, as the models generate table headers only or ill-formatted noisy text, it is infeasible to calculate Model Precision (MP).

MR presents the percentage of ground truth method names that are present in the LLM-generated leaderboard. GPT-4 performs the best, indicating that merely 25.24\% original methods are generated by the LLM on average. Gemini performs poorly in comparison to GPT-4 with only 3.3\% MR. Among the open-source LLMs, Mistral Instruct 7B has a MR of 20.4\%, close to GPT-4. The regular Mistral 7B model performs poorly, attaining less than 1\% recall, indicating that instruction tuning helps the model follow instructions to generate a leaderboard table. The second-best open-source LLM Zephyr Beta 7B has 10.8\% MR. Interestingly, Mistral 7B was the starting point for the Zephyr Beta 7B model, with distilled direct preference optimization applied to learn a chat model with improved intent alignment. Galactica, which is the only LLM trained specifically on scientific texts (research papers, references, \LaTeX, code, DNA sequences, and knowledge bases) performs poorly on both RPLM and LGPLM tasks. 

We present the MP scores for GPT-4, Gemini, Mistral Instruct 7B, and Zephyr Beta 7B, which computes the precision of correctly generated methods in the LLM-generated leaderboard. Mistral Instruct 7B and Zephyr Beta 7B generate long text with multiple repetitive titles, leading to low MP compared to GPT-4. Gemini, on the other hand, generates reasonable-looking method names, but often incorrect, leading to low MP. GPT-4 has the highest precision, with roughly generating 17\% correct method names on average. However, it still indicates that the model hallucinates and generates several method names that are not present in the papers. SP computes the percentage of correctly generated scores with respect to the correctly generated methods in the leaderboard. Gemini which only has 3.38\% method recall, generates 13.87\% exactly correct scores for the 3.38\% correctly generated methods. GPT-4 has a similar SP of 13.06\% for the 25.24\% correctly generated methods. Overall, the model GPT-4 only correctly generates 25.24\% correct methods, and further, the scores for these methods are incorrect 74.76\% times. Upon further inspection, we identify two major challenges in our pipeline design: (i) poor efficiency of the OCR tool (Grobid) in extracting tables data, and (ii) low recall of the BM25 ranker.

\begin{table}[!tbp]
    \centering
    \small{
    \begin{tabular}{lccc}
    \hline
    \rowcolor{Gray}
    \textbf{Model}  & \textbf{MR} & \textbf{MP} & \textbf{SP}\\ \hline
    \multicolumn{4}{c}{7 B Models}      \\ \hline
    Falcon & 0.010 & - & - \\
    Falcon Instruct & 0.002 & - & -  \\
    Galactica & 0.000 & - & - \\
    LLama 2 & 0.024 & - & -  \\
    LLama 2 Chat & 0.077 & - & - \\
    Mistral & 0.351 & - & -  \\
    Mistral Instruct & 20.423 & 5.75 & 2.13 \\
    Vicuna & 0.023 & - & -  \\
    Zephyr Beta & 10.870 & 1.49 & 1.81 \\ \hline
    \multicolumn{4}{c}{13 B Models}      \\ \hline
    LLama 2 & 0.014 & - & - \\
    LLama 2 Chat & 0.020 & - & - \\
    Vicuna & 0.061 & - & -  \\ \hline
    \multicolumn{4}{c}{Closed Models}      \\ \hline
    Gemini Pro & 3.382 & 2.73 & 13.87 \\
    GPT-4 & 25.244 & 17.14 & 13.06\\
    \hline
    \end{tabular}
    }
    \caption{Performance of LLMs on the LGPLM task.}
    \label{tab:lgplm_results}
\end{table}

\section{Related Works}
\label{sec:relatedworks}
\textbf{Information Extraction from Scientific Papers}:
Several works~\citep{viswanathan2021citationie,jain2020scirex,luan2018multi,augenstein2017semeval,luan-etal-2017-scientific} extract dataset, task, method, and metric (DTMM) entities for leaderboards. \citet{bedi2022did} annotate 4.9k references as baselines similar to AP-PCN, and~\citet{kabongo2021automated} curate DTM triples from 4.5k articles. However, these datasets are significantly smaller than our dataset, which contains 1.9M articles, citation and comparison networks from the last 22 years of arXiv, and 70k TDM triples. Parallely, multiple works extract table data from images~\citep{kayal2022tables, tabimghtmlpubtabnet2020}, \LaTeX~\citep{kardas2020axcell,li2020tablebank} and PDFs~\citep{liu2007tableseer}. However, IE works extract entities from the text and tables and don't focus much on leaderboard construction due to several challenges such as entity normalization, determination of metric directionality, merging results, and organizing results into leaderboards.

\noindent\textbf{Automatic Leaderboard Generation}
\citet{hou2019identification} presents two datasets and a framework called TDMS-IE for automatically extracting DTM entities and score information from papers. IBM-TDMS \cite{hou2019identification} and ORKG-TDM~\cite{kabongo2021automated} use an RTE (recognizing textual entailment) task, where given the paper context, the task determines if TDM tuples are entailed, contradicted, or can't be deduced. However,\citet{kabongo2023zero} show that RTE models for DTM identification are not generalizable to new data.
% Models categorize the entailment between the paper text and TDM tuples using BERT~\citep{devlin-etal-2019-bert} and XLNet~\citep{yang2019xlnet}. While IBM-TDMS managed to get an F1 score of 67.8\% on a corpus of 332 papers, ORKG-TDM expanded the work to a bigger dataset of 5361 papers and achieved state-of-the-art performance with F1 scores above 90\% using transformer-based models.
% \citet{bedi2022did} annotate 56,052 references of 2075 papers from the ACL Anthology Reference Corpus~\citep{bird-etal-2008-acl} and mark 4989 references as baselines. 
% They employ citation sentences (citances), similarity of citation context and abstract of the citing paper, reference location, cue words, and citation count of the reference to predict whether a reference is a baseline in the paper. 
% \citet{kabongo2021automated} curate around 4500 scholarly articles and TDM triples from PwC. However, their dataset is significantly smaller than our dataset, which contains 1.9M articles, and citation and comparison networks from the last 22 years of arXiv, in addition to 70k TDM triples. 
% Additionally, recent work by~\citet{kabongo2023zero} showcase that RTE models for TDM identification are not generalizable to new TDM if any is unseen before, indicating that even though the context is given, the models learn classification and not entailment. 
% This was showcased by an $ORKG-TDM_{BERT}$ model whose macro F1 performance dropped from 90.8\% to 26.7\% as the non zero-shot setting was transformed to zero-shot RTE.
In a parallel direction, AxCell~\citet{kardas2020axcell} and~\citet{yang2022telin} present an end-to-end ML pipeline for extracting results from papers.
% with modules for result table detection, table segmentation, abbreviation detection, and linking.~\citet{yang2022telin} follows up in a similar line of work by designing a pipeline for extracting leaderboards from PDFs on the AxCell dataset.
~\citet{singh2019automated} consolidate tables from multiple papers into a graph that illustrates performance improvements. Every individual performance edge is extracted from a table with citations to other papers. 
In contrast, our setting is flexible and realistic, as it starts with a natural language query of <D, T, M>.
% These extractions resemble (noisy) outcomes of `matches' in an incomplete tournament, and several page rank approaches are presented to rank papers. %In contrast, our setting is flexible and starts the leaderboard generation process with a query consisting of a <T, D, M> triple. We formulate the task as a ranking of papers by ranking in a graph constructed for the TDM query, and also as a ranking problem by prompting language models. 
% The dataset can also be utilized as a QA dataset that extracts results for a TQM question.

\section{Conclusion}
\label{sec:conclusion}
We curate two dataset collections, APC and PwC-LDB, to construct \Bench~benchmark for automatic scientific leaderboard generation task. \Bench~largely caters to graph-based rankers and language models for leaderboard generation. We design three tasks, with six configurations to comprehensively evaluate diverse systems for automatic scientific leaderboard generation. 
Across all three tasks, we find that existing models severely lack the capabilities to generate scientific leaderboards leveraging paper texts and paper network datasets. This opens up a new avenue for foundation models to focus on, which also helps the community by reducing the overload of comparing and organizing scientific output by generating leaderboards.
In addition to the automatic leaderboard generation problem, our proposed datasets and \Bench~can also be used in traditional tasks such as citation recommendation and intent identification, impact prediction, novelty assessment in review generation, citation count prediction, and venue recommendation for manuscript submission.
% , etc. AP-PCN dataset provides a stronger citation intent signal than traditional citation networks like AP-CN. PwC-LDB, along with APC, can be used to train models for performance extraction. 
% In contrast to the existing saturated scientific benchmarks like SciDocs~\citep{cohan-etal-2020-specter} and SciRepEval~\citep{singh2022scirepeval}, \Bench~ can be used to evaluate the full-text encoding capabilities of large language models. Further, automatic leaderboard generation has applications in AI assistants for researchers and recommendation of comparison baselines.

\section{Limitations}
\textsc{LEGOBench}~benchmark currently only includes ranking tasks, which we plan to extend to various formats such as score extraction and question-answering. Our performance comparison network (AP-PCN) is currently based on the identification of citation patterns in tables in papers. However, it should be noted that not all tables are result comparison tables. Similarly, it is not necessary that all papers whose results are compared are included in tables, and sometimes, results can be compared in the text alone. Such comparison papers won't be present in our performance comparison network. Our leaderboard dataset was curated from PapersWithCode (PwC), and any papers missing in the PwC or arXiv dataset (APC) were excluded from the ranking task dataset. Our ranking task dataset excluded papers with less than three entries and also removed multiple models from the same paper. Hence, the ranking task dataset is not exhaustive. Additionally, our ranking task only focuses on ranking without verifying if the rankings were based on actual scores presented in the candidate papers. 
% We only present baselines that rank without score verification; however, the score values are present in the dataset, and we urge researchers to use scores for ranking. 
Our PwC-LDB dataset is curated from the PaperWithCode repository. An adversary can add incorrect results, leading to poor performance of good ranking models. Similarly, adversaries can also add incorrect results to existing leaderboards to favor a specific group or an individual organization. Lastly, we rely on PwC to correctly map the scores to the corresponding papers, even if they are reported as baselines in other papers.

\bibliography{anthology,custom}

\appendix
\section{Appendix}
\label{sec:appendix}
\subsection{Exponential Growth in Monthly Paper Submissions on arXiv: 1995-2022}
\label{ssec:app_expgrowth}
\begin{figure}[!htbp]
    \centering
    \includegraphics[width=.9\linewidth]{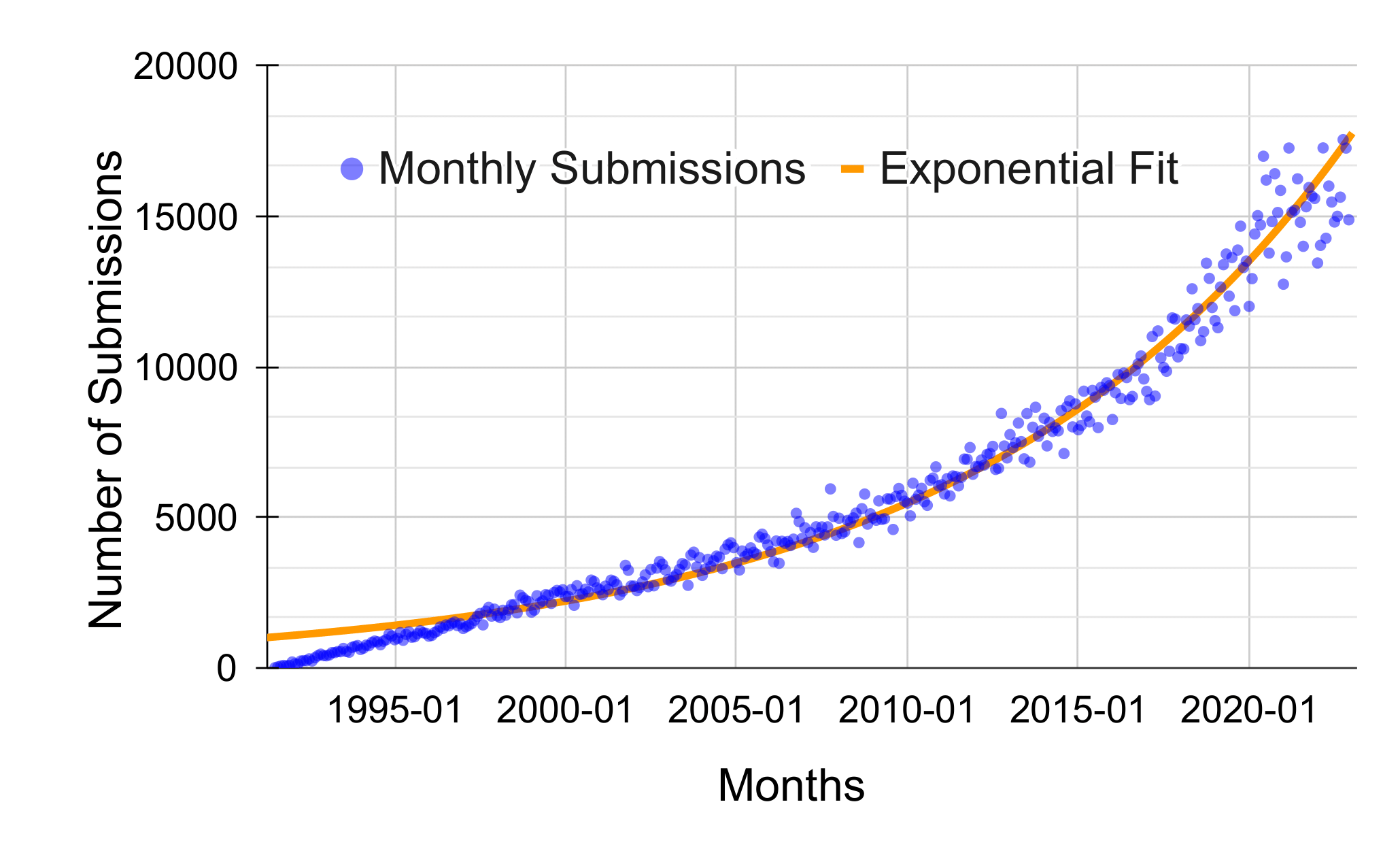}
    \caption{The graph illustrates the exponential growth in the number of papers published monthly on arXiv from 1995 to 2022. This trend showcases the continuous expansion of research and knowledge in the academic community.}
    \label{fig:arXivgrowth}
\end{figure}

Figure~\ref{fig:arXivgrowth} presents monthly submissions to arXiv from 1995 to 2002, indicating the exponential growth in submitted manuscripts.

\subsection{Representative leaderboard taken from PapersWithCode}
\label{ssec:rep_ldb_pwc}
\begin{figure}[!htbp]
    \centering
    \includegraphics[width=.95\linewidth, frame]{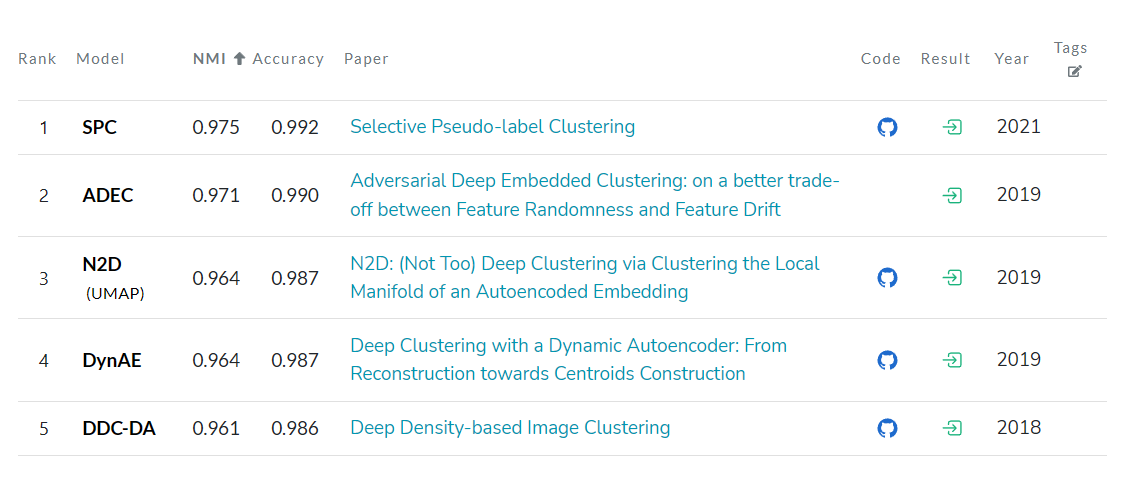}
    \caption{A snapshot of the leaderboards from PwC showcasing top-performing models for Image Clustering on MNIST Dataset and ranked based on NMI (Normalized Mutual Information) metric. Image clustering in the MNIST dataset is the process of grouping similar handwritten digit images and the NMI metric measures how well the clusters align with the actual categories.
    }
    \label{fig:rep_leaderboards}
\end{figure}

We present a snapshot of a leaderboard taken from PapersWithCode in~\Cref{fig:rep_leaderboards}.

% \newpage
\subsection{Dataset Statistics}
\label{ssec:app_networkstats}
The size and the connected components in the citation and performance comparison network are presented in~\Cref{tab:network-graph-info}. The statistics of different dataset modalities and tasks in the dataset are presented in~\Cref{tab:apldb_data_modality} and~\Cref{tab:apldb_task_modality} respectively. We finally also present the statistics of the Citation network and the Performance Comparison network for each domain in~\Cref{tab:arxiv_data_domain}. The citation network for the Physics domain is the densest, while for the Economics citation network is the sparsest. 
\begin{table}[!htbp]
    \centering
    \begin{tabular}{lr}
        \hline
        \rowcolor{Gray}
        \textbf{Citation Network} & \textbf{Property} \\
        \hline
        Directed & Yes \\
        Nodes & 18,325,578 \\
        Edges & 59,890,375 \\
        |$\mathcal{CC}$| & 0.993436 \\
        |$\mathcal{SCC}$| & 0.001794 \\\hline
    \end{tabular}
    \quad
    \begin{tabular}{lr}
        \hline
        \rowcolor{Gray}
        \textbf{Comparison Network} & \textbf{Property} \\
        \hline
        Directed & Yes \\
        Nodes & 280444 \\
        Edges &  309483\\
        |$\mathcal{CC}$| & 0.518481 \\
        |$\mathcal{SCC}$| & 0.000043 \\\hline
    \end{tabular}
    \caption{Statistics of the Citation network and Comparison network. |$\mathcal{CC}$| and |$\mathcal{SCC}$| denote the number of connected components and strongly connected components, respectively.}
    \label{tab:network-graph-info}
\end{table}

% \subsection{Appendix 1.4: Distribution of arXiv Papers by Category: Jan 2000 - July 2022 in Performance Comparison  Network and Citation Network.}
%TABLE 4

\begin{table}[!htbp]
    \centering
    \begin{tabular}{cc}
        \rowcolor{Gray}
        \hline
        Dataset Modality & Frequency \\ \hline
        Images & 498 \\ 
        Texts & 364 \\ 
        Videos & 166 \\ 
        Graphs & 101 \\ 
        Environment & 69 \\ 
        Audio & 41 \\ 
        Lidar & 39 \\ 
        Medical & 34 \\ 
        3d & 26 \\ 
        Point Cloud & 21 \\ 
        RGB-D & 19 \\ 
        Speech & 15 \\ 
        Time Series & 14 \\ 
        Tracking & 12 \\ 
        Tabular & 10 \\ 
        Others & 44 \\ \hline
    \end{tabular}
    \caption{Frequency of various dataset modality of datasets in the AP-LDB dataset. The modality information is taken from PapersWithCode repository.}
    \label{tab:apldb_data_modality}
\end{table}
\quad
\begin{table}[!htbp]
    \centering
    \begin{tabular}{cc}
        \rowcolor{Gray}
        \hline
        Task Category & Frequency \\ \hline
         Computer Vision & 1180 \\
         NLP & 350 \\
         Graphs & 109 \\
         Playing Games & 74 \\
         Miscellaneous & 42 \\
         Medical & 40 \\
         Time Series & 33 \\
         Reasoning & 30 \\
         Speech & 28 \\
         Knowledge Base & 13 \\
         Audio & 12 \\
         Computer Code & 11 \\
         Robots & 7 \\
         Music & 4 \\
         Adversarial & 2 \\
         Others & 146\\ \hline
    \end{tabular}
    \caption{Frequency of various task categories in the AP-LDB dataset. The category information is taken from PapersWithCode repository.}
    \label{tab:apldb_task_modality}
\end{table}

\begin{table*}[!htbp]
    \centering
    \begin{tabular}{p{0.5\columnwidth}|r|rr|rr}
    \hline
    \rowcolor{Gray}
    \cellcolor{Gray} & \cellcolor{Gray} & \multicolumn{2}{c|}{\textbf{PC Network}} & \multicolumn{2}{c}{\textbf{Citation Network}} \\ %\cline{3-6}
    \multirow{-2}{*}{\cellcolor{Gray}\textbf{Domain}} & \multirow{-2}{*}{\cellcolor{Gray}\textbf{Papers}} & \multicolumn{1}{r}{\cellcolor{Gray}\textbf{Nodes}} & \cellcolor{Gray}\textbf{Edges}  & \multicolumn{1}{r}{\cellcolor{Gray}\textbf{Nodes}} & \cellcolor{Gray}\textbf{Edges} \\ \hline
    Computer Science & 618010 & \multicolumn{1}{l}{} 61677 & 18738 & \multicolumn{1}{l}{}  188269 & 2510609  \\ 
    Economics & 5924 & \multicolumn{1}{l}{} 639 & 47& \multicolumn{1}{l}{} 1633 & 10394  \\ 
    Electrical Engineering and Systems Science & 50456 & \multicolumn{1}{l}{} 6864 &375 & \multicolumn{1}{l}{} 14089  & 95145 \\ 
    Mathematics & 530067 & \multicolumn{1}{l}{} 5225 & 20350& \multicolumn{1}{l}{} 171825 & 2068125 \\ 
    Physics & 1123204 & \multicolumn{1}{l}{} 360071 &37952 & \multicolumn{1}{l}{} 360071    & 4296484 \\ 
    Quantitative Biology & 36298 & \multicolumn{1}{l}{} 1313 &700 & \multicolumn{1}{l}{} 11098 & 97310 \\ 
    Quantitative Finance & 14881 & \multicolumn{1}{l}{} 699 & 294& \multicolumn{1}{l}{} 4630 & 41599  \\ 
    Statistics & 157189 & \multicolumn{1}{l}{} 9973 &3485 & \multicolumn{1}{l}{} 50761 & 87695  \\ \hdashline
    \textsc{Total Papers} & 1938693 & \multicolumn{1}{l}{} 613333 & 4644 & \multicolumn{1}{l}{} 107715 & 5419356   \\ \hline
    \end{tabular}
    \caption{Distribution of arXiv papers from Jan 2000-July 2022 in each category. The details of the Performance Comparison  network and the Citation network are listed. This data includes the papers not present on arXiv but present in AP-CN.}
    \label{tab:arxiv_data_domain}
\end{table*}

\subsection{Baseline System Design}
\label{app:baseline_diagrams}
In this section, we present a detailed overview of our baselines for the RPG, RPLM, and LGPLM tasks. For inferencing with LLMs, we use the vLLM library. We had access to a 64 core Intel(R) Xeon(R) Gold 6226R CPU @ 2.90GHz, running Ubuntu 20.04 with 355GB RAM, and one 32GB Nvidia Tesla V100 GPU. We utilized the GPUs for embedding the paper content using the SciBERT, SPECTER, SciNCL, and OAG-BERT, and for LLM inferencing for RPLM and LGPLM tasks. We used NLTK for preprocessing text.

\begin{figure*}[!htbp]
    \centering
    \includegraphics[width=.99\linewidth]{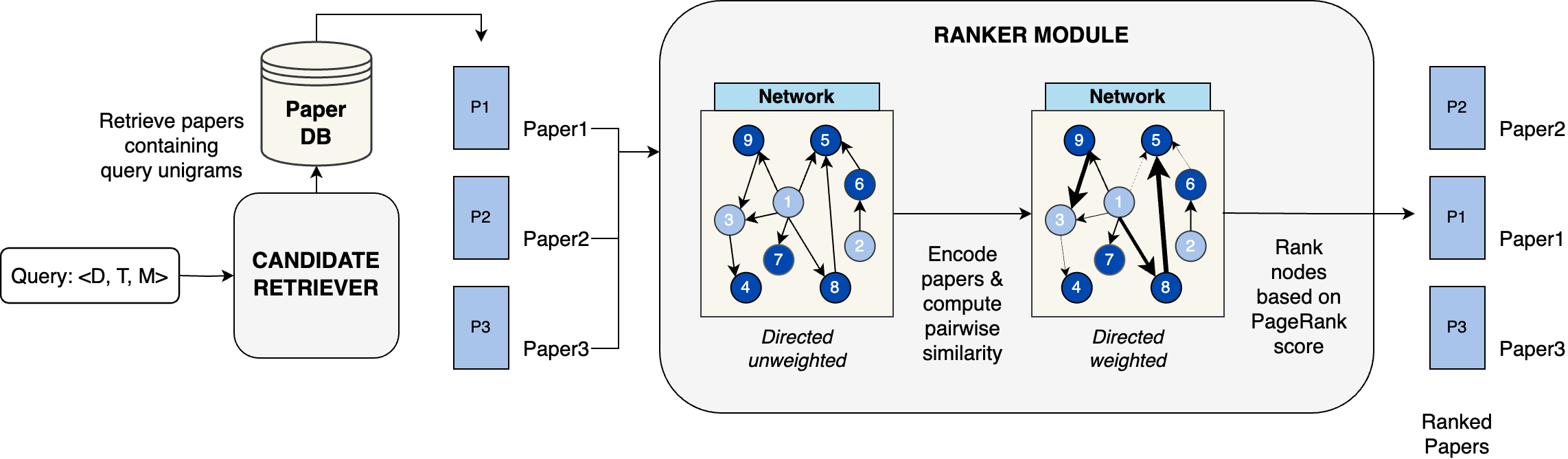}
    \caption{Pipeline for ranking papers with content and graph for leaderboard generation (RPG).}
    \label{fig:rpg}
\end{figure*}

The RPG pipeline is presented in~\Cref{fig:rpg}. It is provided with the natural language query (consisting of D, T, M) and papers dataset and network dataset from the arXiv Papers' Collection. The candidate retriever generates a set of initial candidate papers that report performance on the <D, T, M> triple. The Ranker module leverages the AP-CN or AP-PCN dataset depending on the RPG task configuration and generates a ranking of the papers.

\begin{figure}[!ht]
    \centering
    \includegraphics[width=0.95\linewidth]{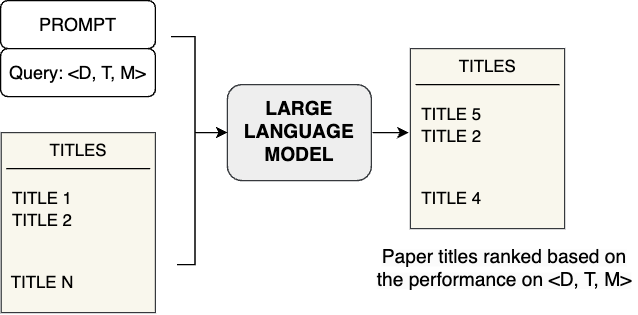}
    \caption{Pipeline for ranking papers by prompting language models (RPLM).}
    \label{fig:rplm}
\end{figure}

RPLM pipeline is straightforward as it involves prompting a language model with the natural language query (<D, T, M>) and a list of paper titles. We provide succinct instructions in the prompt to explain the task to LLM. The pipeline is presented in~\Cref{fig:rplm} and a representative prompt is presented in~\Cref{tab:rplm_prompt}.

\begin{figure*}[!htbp]
    \centering
    \includegraphics[width=.9\linewidth]{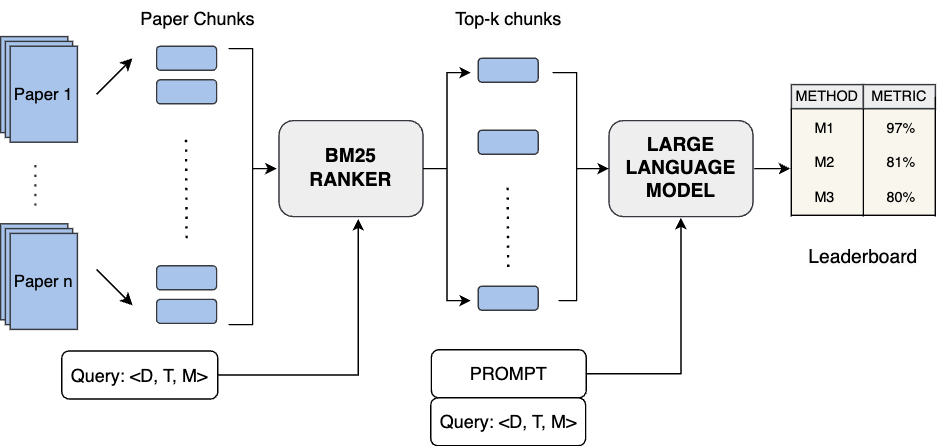}
    \caption{RAG pipeline for leaderboard generation by prompting the language model (LGPLM). Given paper chunks and a query, a BM25 ranker selects the top-10 chunks that are used by an LLM for generating the leaderboard.}
    \label{fig:lgplm}
\end{figure*}

Our baseline for LGPLM is a RAG setup. For the natural language query consisting of <D, T, M> details, we select relevant papers. We split these paper texts into chunks and a BM25 ranker function selects top-10 chunks with respect to the query. The top-10 chunks, query, and an instruction are fed to an LLM, and asked to extract leaderboard entries (methods and scores) from the chunks. The pipeline is presented in~\Cref{fig:lgplm} and a representative prompt is presented in~\Cref{tab:lgplm_prompt}.

We experimented with different prompts and based on empirical observations, we design the prompt to include the following details:
\begin{itemize}[noitemsep,nolistsep]
    \item Task description alongwith the input and output details.
    \item Specification of the domain, machine learning in our case.
    \item Instruction to not generate any additional text other than the specified output.
\end{itemize}

\begin{figure*}[!htbp]
    \begin{minipage}[b]{.45\linewidth}
        \centering
        \small{
        \begin{tabular}{|p{0.9\linewidth}|}
        \hline
        \rowcolor{Gray}
        \textbf{PROMPT:} \\ 
        You are provided with a list of paper titles in the machine learning domain. Your task is to rank them based on their performance on a specific task and dataset using a metric mentioned in the query(the best-performing model should be listed first and the worst should be listed last). Use only the best-performing model proposed in the papers below to compute the ranks. Only include the ranked list of titles in your response and skip any additional text. Query - Rank the performance of the following papers on the <TASK> on dataset <DATASET> using metric <METRIC>. \\
        \textbf{T1:} \dots \\
        \textbf{T2:} \dots \\
        \textbf{T3:} \dots \\
        \textbf{T4:} \dots \\ \hline
        \end{tabular}}
        \caption{Prompt Template for Ranking by Prompting Language Models [RPLM]. <TASK>, <DATASET>, and <METRIC> are replaced by appropriate T, D, and M values in the prompt.}
        \label{tab:rplm_prompt}
    \end{minipage}\hfill
    \begin{minipage}[b]{.45\linewidth}
        \centering
        \small{
        \begin{tabular}{|p{0.9\linewidth}|}
        \hline
        \rowcolor{Gray}
        \textbf{PROMPT:} \\ 
        Excerpts: .... \\
        You are provided with a dataset, task, and metric. You need to create a leaderboard which lists the performance of various methods on the provided dataset and task using the provided metric. Excerpts from research papers are provided above which report the performance of methods on these task, dataset and metric. Extract the performance from the excerpt to create the leaderboard. The output should be a single table listing each method and performance only. Do not include any explanation or additional text in the output, only include method name and performance scores. Query - List the performance scores of various methods on the <DATASET> dataset on the <TASK> task using metric <METRIC>. \\ \hline
        \end{tabular}}
        \caption{Prompt Template for Leaderboard Generation by Prompting Language Models [LGPLM]. <TASK>, <DATASET>, and <METRIC> are replaced by appropriate T, D, and M values in the prompt.}
        \label{tab:lgplm_prompt}
    \end{minipage}
\end{figure*}

\subsection{Acronyms and Abbreviations used in the paper}
We present acronyms and abbreviations used in the paper in~\Cref{app:acronyms}.
\begin{table*}[!htbp]
\begin{tabular}{p{0.2\linewidth}l} \hline
\rowcolor{Gray}
\multicolumn{2}{c}{Dataset}   \\ \hline
APC               & arXiv Papers' Collection      \\
AP-FT             & arXiv Papers' Full Text      \\
AP-TABS           & arXiv Papers' Title \& Abstract \\
AP-CN             & arXiv Papers' Citation Network \\
AP-PCN            & arXiv Papers' Performance Comparison Network \\
AP-LDB            & arXiv Papers' Leaderboard    \\
PwC-LDB           & Papers with Code Leaderboard \\ \hline
\rowcolor{Gray}
\multicolumn{2}{c}{Task} \\ \hline
RPG  & Ranking Papers based on Content and Graph   \\
RPG[CN-TABS]  & Ranking Papers in the Citation Network with Titles and Abstracts \\
RPG[CN-FT]    & Ranking Papers in the Citation Network with Full Text \\
RPG[PCN-TABS] & Ranking Papers in the Performance Comparison Network with Titles and Abstracts \\
RPG[PCN-FT]   & Ranking Papers in the Performance Comparison Network with Full Text \\
RPLM & Ranking Papers by Prompting Language Models  \\
LGPLM & Leaderboard Entries Generation by Prompting Language Models \\ \hline
\rowcolor{Gray}
\multicolumn{2}{c}{Metrics}                  \\ \hline
BEM               & Binary Exact Match       \\
CIS               & Complete Inclusion Score \\
CP                & Concordant Pairs         \\
KTau              & Kendall's Tau            \\
MR                & Method Recall            \\
MP                & Method Precision         \\
SP                & Score Precision          \\ \hline
\rowcolor{Gray}
\multicolumn{2}{c}{\textbf{Misc}}                     \\ \hline
DTMA              & Dataset, Task, Metric, Algorithm/Method/Model               \\ 
FT                & Full Text (in the context of research papers) \\
PwC               & Papers with Code         \\
TABS              & Title \& Abstract (in the context of research papers)  \\ \hline
\end{tabular}
\caption{List of acronyms and abbreviations used in the paper.}
\label{app:acronyms}
\end{table*}

\end{document}